\documentclass[10pt,conference,a4paper]{IEEEtran}
\usepackage{booktabs}
\usepackage{graphicx}

\usepackage{xparse}
\usepackage{amsmath}
\usepackage{amssymb}
\usepackage{color}

\usepackage[font=small]{caption}

\usepackage{notation_memmesheimer}
\usepackage{hyperref}
\usepackage{lipsum}

\usepackage{times}
\usepackage{epsfig}
\usepackage{stmaryrd,pgfplots,multirow,hhline,tikz}

\def\andothers{et al.\ }

\ExplSyntaxOn
\NewDocumentCommand{\storereviewer}{mm}
  {
   \bcp_store_data:nn { #1 } { #2 }
  }

\DeclareExpandableDocumentCommand{\getreviewer}{O{1}m}
 {
  \bcp_get_data:nn { #1 } { #2 }
 }

\cs_new_protected:Npn \bcp_store_data:nn #1 #2
 {
  \seq_if_exist:cF { l_bcp_data_#1_seq } { \seq_new:c { l_bcp_data_#1_seq } }
  \seq_set_split:cnn { l_bcp_data_#1_seq } { } { #2 }
 }
\cs_generate_variant:Nn \seq_set_split:Nnn { c }

\cs_new:Npn \bcp_get_data:nn #1 #2
 {
  \seq_item:cn { l_bcp_data_#2_seq } { #1 }
 }
\ExplSyntaxOff

\usepackage{pgfkeys}
\pgfkeys{
 /points array/.is family, /points array,
 .unknown/.style = {\pgfkeyscurrentname/.initial = #1},
}

\newcommand\resultsstorage[1]{\pgfkeys{/points array, #1}}

\resultsstorage{apsr20 = 29.1, apsr40 = 34.8, apsr60 = 39.2, apsr80 = 42.8, apsr100 = 45.3,
                sldml20 = 36.7, sldml40 = 42.4, sldml60 = 49.0, sldml80 = 46.4, sldml100 = 50.9,
                simitate_22_4=100.0, simitate_18_4=98.8, simitate_14_4=99.4, simitate_10_4=93.1, 
                simitate_18_8 = 76.7, simitate_14_12=61.2, simitate_10_16=56.6,
                utdmhad_inter_joint_wrist_leg=80.4, utdmhad_inter_joint_leg_wrist=28.3,
                utdmhad_inter_joint_wrist_leg_eq_class_dist=18.8, utdmhad_inter_joint_leg_wrist_eq_class_dist=59.7,
                utdmhad_inter_sensor_skeleton_inertial_signals=35.5,
                utdmhad_inter_sensor_inertial_skeleton_signals=40.5,
                utdmhad_inter_sensor_skeleton_inertial_sldml=23.1,
                utdmhad_inter_sensor_inertial_skeleton_sldml=40.5,
                utdmhad_23_4_skeleton=92.7,utdmhad_19_8_skeleton=74.8,utdmhad_15_12_skeleton=81.1,utdmhad_11_16_skeleton=77.7,utdmhad_7_20_skeleton=57.2,utdmhad_3_24_skeleton=59.4,
                utdmhad_23_4_inertial=81.3,utdmhad_19_8_inertial=74.0,utdmhad_15_12_inertial=63.5,utdmhad_11_16_inertial=43.3,utdmhad_7_20_inertial=41.3,utdmhad_3_24_inertial=29.2,
                utdmhad_23_4_fused=90.2,utdmhad_19_8_fused=76.0,utdmhad_15_12_fused=78.7,utdmhad_11_16_fused=69.4,utdmhad_7_20_fused=65.0,utdmhad_3_24_fused=44.7,
                }
\usepackage[detect-none]{siunitx}
\pgfplotsset{compat=1.14}

\begin{document}
%

\title{Adaptive Local-Component-aware Graph Convolutional Network for One-shot Skeleton-based Action Recognition}


\author{\IEEEauthorblockN{Anqi Zhu, Qiuhong Ke, Mingming Gong, James Bailey}
\IEEEauthorblockA{The University of Melbourne, Australia\\
\texttt{azzh1@student.unimelb.edu.au, \{qiuhong.ke, mingming.gong, baileyj\}@unimelb.edu.au}}
}

\maketitle

\begin{abstract}
Skeleton-based action recognition receives increasing attention because the skeleton representations reduce the amount of training data by eliminating visual information irrelevant to actions. To further improve the sample efficiency, meta-learning-based one-shot learning solutions were developed for skeleton-based action recognition. These methods find the nearest neighbor according to the similarity between instance-level global average embedding. However, such measurement holds unstable representativity due to inadequate generalized learning on local invariant and noisy features, while intuitively, more fine-grained recognition usually relies on determining key local body movements. To address this limitation, we present the Adaptive Local-Component-aware Graph Convolutional Network, which replaces the comparison metric with a focused sum of similarity measurements on aligned local embedding of action-critical spatial/temporal segments. Comprehensive one-shot experiments on the public benchmark of \textit{NTU-RGB+D 120} indicate that our method provides a stronger representation than the global embedding and helps our model reach state-of-the-art. 
\end{abstract}


%
\IEEEpeerreviewmaketitle

\section{Introduction}
\label{sec:intro}
Action recognition is one of the computer vision problems that have great practical importance, widely used in modern applications such as auto surveillance systems \cite{paper1}, video retrieval \cite{paper2}, etc. The past research on this is majorly for RGB-based inputs due to their wide accessibility. Yet, pixelated inputs have a high risk of information over-richness, which could easily make a model over-volume and confused by the task-irrelevant background, brightness, and color changes \cite{data-modality}. A 3-D Skeleton Sequence becomes one of the strong input alternatives \cite{paper3} because it only records 3-d body joint movements along temporal evolution, and is easily obtainable from RGB-based inputs by a pre-trained pose estimation algorithm \cite{6,3,5}.
\begin{figure}[t]
\centering
\includegraphics[width=\columnwidth]{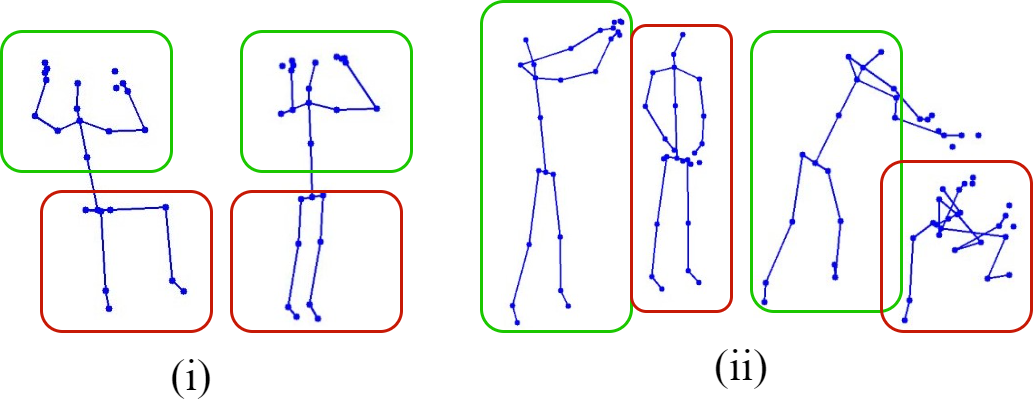}
\caption{Frame-shot examples from the classes ``Putting on a cap" (i) and ``Hitting another person with something" (ii) in \textit{NTU-RGB+D 120} \cite{3}. To achieve a fast adaptation to valid recognition, we expect to prioritize the similarity discrimination for the spatial-temporal features from the movements in the green blocks, and suppress the noisy features from the red blocks.}
\label{fig:pa-diff}
\end{figure}

Using deep learning models and vast annotated training samples, the existing skeleton-based action recognition solutions implemented highly-accurate classification for pre-known activities. But the study on extending prediction to foreign classes is still at a beginning stage. Known as few-shot learning (FSL), it is an active research topic to realize fast adaptation to the classification tasks for new classes with a handful of supervised examples. Especially, the situation is called a one-shot learning (OSL) problem when only one example is available for each new class. Its solutions would greatly improve the models' practicality for real-world missions by reducing the dependency on data-intensive training environments and enabling one-time/rare-case learning for which other supervised method is not available \cite{fewshot-survey}. 

Among current skeleton-based solutions that support OSL, early studies first transformed skeleton sequences to signal images to solve the problem as unified image-based classification \cite{skl-dml,sl-dml}. But the transformation deforms the original skeletal structure and inevitably causes inefficient feature extraction and information loss on the spatial connectivities between parallel neighboring body joints. A more preferable way is to let a model achieve a metric-based OSL by comparing the similarity between inputs' projection on a common embedding space that retains more integral descriptions by native skeleton-based backbones. The solutions in \cite{3, tcn} respectively achieved different implementations for this but their results are still inferior to the signal-based solutions. During their classification process, encoded features are averagely pooled to generate a single global embedding as an input's representation for similarity comparison. If having adequate supervised training, such representation could self-accumulate enough distinction learning on the latent action-invariant features at a global scope. However, the directly learnable references for few-shot classes are too sparse. Thus the generality of their global embedding could be easily disturbed by the local bias features, and fails to robustly focus on the necessary fine-grained invariant features for refined recognition. On the other hand, we observe that specific local features divided under body-part-based partitions or consecutive time sections intuitively separate action-critical/irrelevant patterns for valid/invalid recognition clues. As shown in Fig. \ref{fig:pa-diff}, while generalizing an abstract global pattern across all body joints and frames is difficult, a linear combination of aligned discrimination on the patterns in the green boxes should help quickly determine a good recognition for the given inputs. Similarly, the patterns in the red boxes are apparently non-action relevant and the measurements on their embedding (or the global embedding that include them) become  learning noise that should be directly suppressed to reduce representation bias. 

In this paper, we propose Adaptive Local-Component-aware Graph Convolutional Network (ALCA-GCN) as the first metric-based method to rely on local embedding distancing as the main determinant factors for one-shot skeleton-based action recognition. It decomposes the instance-level similarity matching to a selective sum of body-part-based local embedding measurements under ordered temporal sections. To achieve this, we start with an encoding backbone that extracts hierarchical spatial-temporal features for body-part-level pattern and skeleton-level contexts. Our embedding function then performs averagely pooling over the encoded features under segmented local areas. This generates the independent representation for the comparing unit of each body part under different time sections. When calculating the total similarity, our model sequentially aggregates the distance of each aligned unit embedding between the given support and query inputs, and applies an adaptive emphasis/suppression for the decision impact from the measurements on action-critical/noisy units. We evaluate our solution on \textit{NTU-RGB+D 120} \cite{3} and use the official OSL testing protocol to compare with all previous related papers \cite{3,skl-dml,sl-dml,tcn}. The result proves that our model achieves state-of-the-art performance. Concretely, our contributions are:  
\begin{itemize}
\item We propose ALCA-GCN as a novel metric-based OSL solution for skeleton-based action recognition. It models an action as a matrix of comparable local embedding on spatial and temporal dimensions. The matrix element is divided according to body parts and average temporal sections. 
\item ALCA-GCN determines the total similarity between two skeleton sequences by a selective sum of embedding distances between all aligned local comparing units in spatial and temporal dimensions.  
\item During the similarity aggregation, ALCA-GCN self-learns the emphasis/suppression against the comparison importance from action-critical/irrelevant units. The model presents better results than the prior art under an extensive one-shot learning experiment setup using \textit{NTU-RGB+D 120}.
\end{itemize}

\section{Related Work}
\label{sec:related}
\subsection{RGB-based Image/Video FSL} 
Being the primary experimental ground for FSL, many solutions have been developed for image classification, systematically divided into data-based, model-based, and meta-learning-based approaches \cite{fewshot-survey}. Especially, as one of the meta-learning-based methods, metric-based learning is sharply focused due to its simple structure and flexible component scalability. In 2017, Snell \andothers \cite{protoNet} proposed one of its major frameworks, which classifies according to the nearest Euclidean distances between queries and class prototypes in a well-generalized common embedding space. To align the training feature distribution to true few-shot testing tasks, Vinyals \andothers \cite{matching-net} devised the episodic learning strategy which trains a model by a multi-tasking learning procedure. During a training epoch, each sub-task simulates the same N-way-K-shot setting from the testing conditions (i.e. having K reference instances for N candidate classes). Based on the above learning framework, various solutions are further designed to improve the few-shot generalization ability for each component, including enriching embedding features with external knowledge (e.g. semantic \cite{paper12}), devising local-descriptor-based similarity matching \cite{localdesc, deepemd}, empowering learning ability to metric functions \cite{aml,tadam}, etc. 

FSL for RGB-video-based action recognition requires additional learning on the temporal dimension. Tan and Yang \cite{paper23} first regarded this as a variant for image classification by compressing the input videos into static dynamic images. It was not until the breakthrough of deep volumetric extraction tailored for video features \cite{3dcnn, r2+1d}, that many papers \cite{protogan, many-way-few-shot,  ar-permutation-alignment,ar-temporal-alignment} started to adopt the new backbones with common FSL frameworks. The difficulty of generalizing a video-level embedding space comes from the temporal dimension expansion and thus the exponential increase of sample variance and backbone volume \cite{paper6}. To mitigate the deep learning pressure, recent papers look for alternatives that could figure total similarities based on extracted local features in non-parametric ways. In 2021, Ben-Ari \andothers \cite{taen} presented a metric-based solution in which total similarities between queries and generated class prototypes are measured by the sum of spatial feature differences in averagely divided time sections. Cao \andothers \cite{paper25} applied Dynamic Temporal Warping (DTW) to orderly aggregate the similarities between the closest spatial embedding matches for every frame in two videos. While the local feature comparison on the temporal dimension is decomposable according to the action evolution order, pixelated inputs are hard to define meaningful and stationary geographical partitions for spatial local features under a frame. On the other hand, a performer's skeletal description maintains an invariant graph structure along the time, persisting with explicit component meaning for each body part.

\subsection{Skeleton-based Action Recognition}
Research on the deep feature extraction of skeleton sequences is steadily developed in the past few years. Regarding action recognition as a temporal modeling problem, early solutions adopted RNN/LSTM-based extraction by sequentially feeding frame-level body joint inputs and predicting according to the accumulated learning status in the last frame \cite{rnn-skl-fewshot,st-lstm}. \cite{2-stream-rnn} refined spatial feature encoding with another parallel RNN model which establishes sequential connectivities among body joints according to a pre-defined traversing path along the whole skeleton. Since recursive connections are not competitive for their spatial modeling ability, \cite{cnn-skl-fewshot} replaced the encoder with a CNN that parallelly convolutes the features from linearly arranged adjacent body joints and frames. In 2018, research reveals that the kinetic dependencies in body joints' native multi-neighbor connections transmit more integral and abundant spatial information. Yan \andothers \cite{2} proposed a Spatial-Temporal Graph Convolutional Network (ST-GCN), which supports temporal and spatial feature convolution based on an adaptive multi-neighbor sampling scheme. The original scheme convolutes features from body joints' relative distances to the body center. It consists of a set of adjacency matrices and is scalable for customizing more relations such as bone connections \cite{2s-gcn} and directional kinetic transmission \cite{dg-gcn}. We benefit from this advantage and design a hierarchical convolution to obtain more refined features for local partition representation. 

\subsection{FSL in Skeleton-based Action Recognition}
In the existing solutions, the topic is first tackled by Liu \andothers \cite{3}, who extracted skeletal features by an ST-LSTM backbone \cite{st-lstm} and adopted a Euclidean-distance-based similarity comparison on a sharing global embedding space. To compensate for the generalization insufficiency from limited samples, it imports external prior knowledge on the semantic relations between body joint names and instance labels. The relevance scores re-assign the contribution weights of body-joint-level features toward the global embedding to emphasize class-related body joint feature learning. \cite{tcn} convolutionally extracted temporal features from the normalized body-joint coordinate average for each frame, and regarded the features from the last frame as the final instance-level representation for similarity comparison. \cite{skl-dml,sl-dml} devised a pre-processing module that transforms skeleton sequences into signal images by lining up body joints as rows and frames as columns. The 3-d coordinate for each joint under every frame becomes a 3-channel pixel value. The compression deforms the parallel connectivity between body joints in the original skeleton structures. Different from the previous solutions, our solution doesn't rely on external knowledge and format transformation, but decomposes the comparing metric to an adaptive sum of local measurements. Thus it could self-emphasize the learning distinction in action-critical local areas. 

\section{Approach}
\label{sec:method}
\begin{figure*}[ht]

\centering
\includegraphics[width=\textwidth]{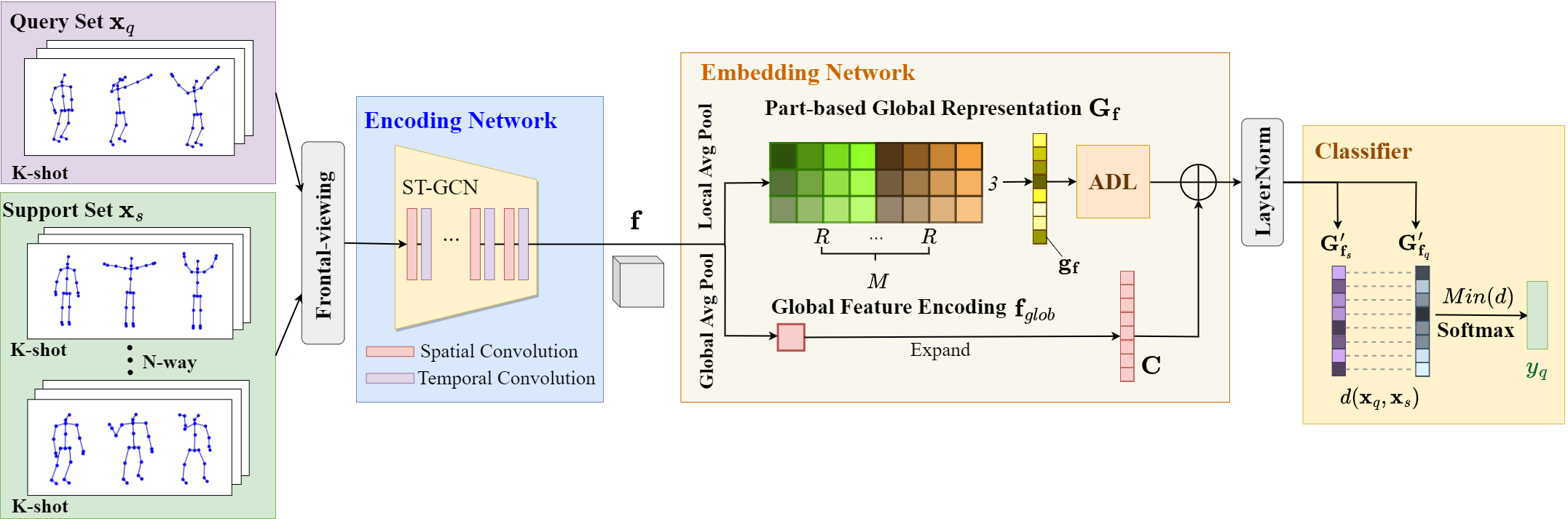}
\caption{The overview of ALCA-GCN. Each input is first pre-processed as frontally-viewed to remove viewpoint variances. The encoder $F$ applies two types of convolution on an input $\mathbf{x}$ for body-part-level surrounding and skeleton-level contextual features. The embedding network then partially pools the encoded feature $\mathbf{f}$ to generate the embedding of comparing unit $\mathbf{g}_\mathbf{f}$ for $R$ body parts under 3 temporal sections per each skeleton. Concatenating all $\mathbf{g}_\mathbf{f}$ together forms the complete global representation $\mathbf{G}_\mathbf{f}$ for describing $\mathbf{x}$. The following ADL module highlights the contextual impact from action-critical units based on an self-attention mechanism, and the instance-level global constraint $\mathbf{f}_\mathit{glob}$ is aggregated to each modified unit to impose intra-class clustering. Eventually, the total similarity between a support sample $\mathbf{x_s}$ and a query sample $\mathbf{x_q}$ is determined by the measurement on their representation $\mathbf{G}'_{\mathbf{f}_q}$ and $\mathbf{G}'_{\mathbf{f}_s}$, which is the sum of Euclidean distances between all aligned comparing units.}
\label{fig:ALCA-GCN}
\end{figure*}

Fig. \ref{fig:ALCA-GCN} presents the architecture of ALCA-GCN. Our method follows the basic framework of metric-based solutions, consisting of a feature encoding backbone, an embedder for modeling the representation matrix, and a linear-metric-based classifier. To remove viewpoint variances, we unite all inputs to a frontal viewing angle before the encoding (see Section \ref{subsec:implements}). We adopt an ST-GCN \cite{2} network $F$ as the prototype backbone because it allows feature convolution over each body joint's neighbors following the original human body (sub-)structure. Maintaining its original skeleton-level convolution, we additionally devise independent kernel matrices for body-part-level convolution. So that the modified $F$ captures each body joint's surrounding contextual features under its belonging body part and its relative global position features at the skeleton scope. We use $F$ to obtain the total feature $\mathbf{f}$ from a pre-processed input $\mathbf{x}$ and pool it as a part-based global representation $\mathbf{G}_\mathbf{f}$, which contains a group of local embedding $\mathbf{g}_\mathbf{f}$ for 4 body parts (head, hands, torso and legs) under 3 temporal sections. They are regarded as the basic comparing units for the independent local similarity measurements. To enhance/suppress the feature distinction learning on action-critical/irrelevant sub-areas, an Adaptive Dependency Learning (ADL) module is attached behind $\mathbf{G}_\mathbf{f}$ to adaptively adjust each unit's content influence for the later similarity matching. An average global embedding is also aggregated into each unit as an instance-level constraint to impose the clustering consistency for intra-class sample units. Finally, trained with the total Euclidean distance loss between the aligned embedding units of the given query and its belonging class support example, the model learns to classify with a strong reliance on the similarity of action-critical local patterns. The differences between noisy units are more amended by supplementing their original embedding with other high-attention contexts. The rest of the section further describes the details for each model component.      

\begin{figure}[ht]
\centering
\includegraphics[width=\columnwidth]{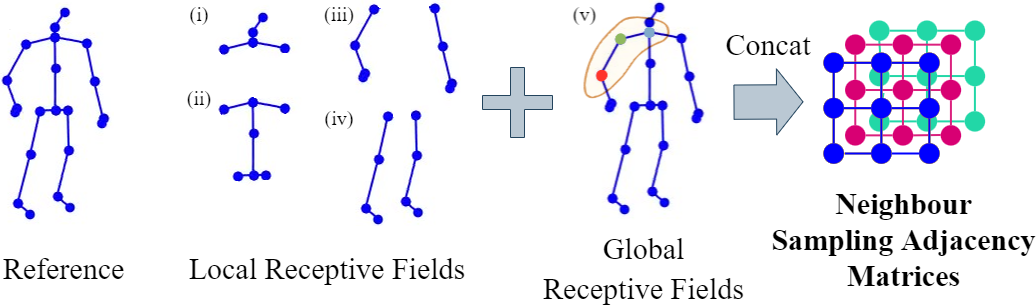}
\caption{Neighbor sampling scheme for spatial convolution in ALCA-GCN. Local receptive fields cover the feature convolution on 4 body-part sub-graphs, including (i) head, (ii) torso, (iii) hands, and (iv) legs. The limbs are symmetrically grouped to avoid handedness disparities. The global receptive field (v) from \cite{2} convolutes on every joint’s root, centrifugal, and centripetal neighbors (figured by their global distances to the body center). }
\label{fig:partition-scheme}
\end{figure}

\subsection{Encoding architecture}
\label{ssec:encoding}
We denote an input skeleton sequence as $\mathbf{x}_{\mathit{orig}} \in \mathbb{R}^{C \times T \times U \times M}$, where $M$ refers to the number of performers, $U$ refers to the number of body joints for one performer, $T$ refers to the number of frames, and $C$ refers to the number of body joint coordinate dimension (usually is 3). Obtaining the adjusted input $\mathbf{x} \in \mathbb{R}^{C \times T \times U \times M}$ from the frontal-viewing pre-process, $F$ first encodes it as $M$ individual sequences $\mathbf{x}' \in \mathbb{R}^{C \times T \times U}$ and later concatenates them back before feeding them to the embedding network.

In the original ST-GCN \cite{2}, $\mathbf{x}'$ is represented as a graph of $\{\mathcal{V}, \mathcal{E}\}$ that includes every body joint from all frames. Concretely, $\mathcal{V} = \{v | v \in \mathbb{R}^{C}\}$ and $|\mathcal{V}| = U \times T$. $\mathcal{E}$ consists of the physical and temporal connections as $\mathcal{E} = \{(v_i, v_j) | (v_i, v_j)\in \mathcal{H}\} \cup \{(v_i^t, v_i^{t+1}) | t\in (0, T)\}$ where $\mathcal{H}$ includes all naturally adjacent body joint pairs in each frame, and $(v_i^t, v_i^{t+1})$ refers to the temporal pair connection of the same body joint $v_i$ between the adjacent frame $t$ and $t+1$. Feature extraction is applied via iterative spatial and temporal graph convolutions. A spatial convolution layer contains two groups of learnable matrices  $\{\mathbf{W}\}_k^{K_S}$ and $\{\mathbf{E}\}_k^{K_S}$. Given an input $\mathbf{f}'\in \mathbb{R}^{C'\times T'\times U}$ from a previous or input layer, for its spatial sub-feature $\mathbf{f}'_{\mathit{in}}\in \mathbb{R}^{C'\times U}$ at each $t'\in T'$, its convolution is calculated as:
\begin{align} \label{eq:st-gcn}
\mathbf{f}'_{\mathit{out}} &= \sum\nolimits_k^{K_\mathcal{S}}\mathbf{W}_k(\mathbf{f}'_{\mathit{in}}\times(\mathbf{A}_k\odot \mathbf{E}_k))\ ,
\\
\mathbf{A}_k &= \mathbf{\Lambda}_k^{-\frac{1}{2}}\bar{\mathbf{A}}_k\mathbf{\Lambda}_k^{-\frac{1}{2}} \label{eq2}\ ,
\\
\mathbf{\Lambda}_k^{mn} &= \begin{cases}
    \sum\nolimits_j^{\mathcal{B}_k^i}(\bar{\mathbf{A}}_k^{mj}),& \text{if } m = n\\
    0,              & \text{otherwise}
\end{cases}\ ,
\end{align}
where $K_\mathcal{S} = L\times R$ is the total number of applied convolutions. $\mathbf{W}_k$ is the $k$-th convolution matrix with a shape of $C''\times C'\times 1\times 1$, where $C''$ is the layer output dimension. $\mathbf{E}_k$ is a $U\times U$ matrix for re-weighting the neighbor features filtered by $\mathbf{A}_k$. $\odot$ is a dot product operation. $\mathbf{\Lambda}_k$ is the degree matrix of $\bar{\mathbf{A}}_k$ to apply degree normalization in Equation (\ref{eq2}). $\bar{\mathbf{A}}_k$ is a pre-defined $U\times U$ adjacency matrix for convolution $k$, where $\bar{\mathbf{A}}_k^{ij}$ indicates whether $v_j$ belongs to the convoluting area $\mathcal{B}_k^i$ for $v_i$. ST-GCN regards a skeleton as a single complicated graph that spreads around a body center joint \cite{2}. By measuring the distances to the spine, it categorizes each joint's physical neighbors into three global relation types $R$, known as Centrifugal, Centripetal and Root (i.e. itself). To apply a one-kernel convolution, $\bar{\mathbf{A}}_r$ records every joint's adjacency status of whether it is a neighbor of type $r$ ($r\in R$) for each body joint. The convolution for each joint is then sampled from its corresponding $R$ neighbors orderly filtered by the matrix multiplication with $\{\bar{\mathbf{A}}_r | r\in R\}$.   

Since our model focuses on the class-level representability under local sub-areas (body parts), except for the joint features from the global position relations to the body center, we also value each joint's relative surrounding features under its belonging body part. We get inspiration from \cite{1} and devise extra $\bar{\mathbf{A}}$ to respectively filter a body joint's neighbor features from all local connections under its belonging body part (see Fig. \ref{fig:partition-scheme}). For a certain body part $\mathcal{P}_r$, $\bar{\mathbf{A}}_r$ filters the convoluting area $\mathcal{B}_r^i$ of $v_i$ by:
\begin{align} \label{eq:partition}
\mathcal{B}_r^i &= \{v_j | \mathbf{d}(v_i, v_j)\leq 1, v_i, v_j \in \mathcal{V}_{\mathcal{P}_r}\}\ ,\\
\bar{\mathbf{A}}_r^{ij} &= 
\begin{cases}
1,  & \text{if $v_j \in \mathcal{B}_r^i$} \\
0, & \text{otherwise}
\end{cases}\ ,
\end{align}
where $\mathbf{d}(v_i, v_j)$ denotes the minimum path between $v_j$ and $v_i$. $\mathcal{V}_{\mathcal{P}_r}$ is the set of body joints included in $\mathcal{P}_r$. Edging joints between any two adjacent body parts are overlappingly included in both partitions, so that all $\bar{\mathbf{A}}_r$ for local sampling cover every natural skeletal connection. A spatial convolution layer eventually applies $L$-kernel groups of convolution on 4 body-part sub-graphs and 1 global skeleton graph. Thus there would be overall $K_\mathcal{S} = L \times 5$ parallel convoluting operations, and the weights for each convolution are learned with its own $\mathbf{W_k}$ and $\mathbf{E_k}$. 

For a temporal convolution layer, given an input $\mathbf{f}'$ from its previous spatial layer, we remain a $3 \times 1$ convolution \cite{2} on its temporal sub-feature $\mathbf{f}'_{\mathit{in}}\in \mathbb{R}^{C'\times T'}$ at each $v_i\in \mathcal{V}$. The convoluting area for $v_i^{t'}$ is the sub-features of $v_i$ at $t'-1$ and $t'+1$. 

At the end, $F$ outputs $\mathbf{f} = F(\mathbf{x}) \in \mathbb{R}^{d_{\mathit{feat}}\times T_{\mathit{feat}}\times U\times M}$ after concatenating back $M$ performers' features. $d_{\mathit{feat}}$ and $T_{\mathit{feat}}$ are the encoding sizes on spatial and temporal dimensions for each body joint of every performer. 

\subsection{Part-based Global Representation} \label{subsec:global-repres}
Having $\mathbf{f}$, we apply a segmented mean pooling on the corresponding body joints and temporal dimensions to get the local embedding $\mathbf{g}_{\mathcal{P}_{ri}^m}$ for each body part $\mathcal{P}_{r}$ of performer $m$ at a temporal section $i$. Beyond the same body part partitioning in Fig. \ref{fig:partition-scheme}, we averagely divide 3 temporal sections, known as the starting, middle, and ending phases. Therefore:
\begin{align} 
\label{eq:graph-tensor}
\mathbf{g}_{\mathcal{P}_{ri}^m} &= \frac{1}{|\mathcal{V}_{\mathcal{P}_r}||T_i|}\sum\nolimits_v^{\mathcal{V}_{\mathcal{P}_r}}\sum\nolimits_t^{T_i} \mathbf{f}_{vt}^m\ ,\\
T_i &= \{t|t\in ((i-1)\times T_{\mathit{div}}, i\times T_{\mathit{div}}]\}\ ,
\end{align}
where $R = 4$ is the number of partitioned body parts for a performer, and $T_{\mathit{div}} = T_{\mathit{feat}} / 3$ is the length of each temporal section. All $\mathbf{g}_{\mathcal{P}_{ri}^m}$ are then concatenated to generate the matrix $\mathbf{G}_\mathbf{f}$ as the instance-level global representation for $\mathbf{x}$. 

To obtain the similarity metric between two inputs, we regard each $\mathbf{g}_{\mathcal{P}_{ri}^m}$ as the embedding of a local comparing unit, and successively aggregate the Euclidean distances between each aligned unit in the two object sequences. In a one-shot learning scenario and if using the episodic learning algorithm \cite{matching-net}, for each epoch, the model meta-trains from a batch of sub-tasks randomly sampled from the auxiliary set. Each sub-task has the same N-way-1-shot setting consistent with the testing task. Having an incoming training/testing query input $\mathbf{x}_q$ and some support instances $\{(\mathbf{x}_{s_1}, s_1), (\mathbf{x}_{s_2}, s_2), ..., (\mathbf{x}_{s_P}, s_P)\}$ for candidate classes $s_1, s_2, ..., s_P$, the classification of $\mathbf{x}_q$ is the same category as its most similar support instance $\mathbf{x}_{s_{\mathit{min}}}$ according to the comparing metric. In other words, the model predicts the probability distribution of $\mathbf{x}_q$ belonging to class $s_n$ via:
\begin{align} \label{eq:distance}
p_\phi(y_q=s_n|\mathbf{x}_q) &= \frac{\exp(-d(\mathbf{x}_q, \mathbf{x}_{s_n}))}{\sum_{p=1}^P\exp(-d(\mathbf{x}_q, \mathbf{x}_{s_{p}}))}\ ,\\
    \begin{split}
        d(\mathbf{x}_q, \mathbf{x}_{s_p}) &= d\langle\mathbf{G}_{\mathbf{f}_q}, \mathbf{G}_{\mathbf{f}_{s_p}}\rangle \\
        &= \sum\nolimits_j^{3\times R\times M}\lVert\mathbf{g}_{\mathbf{f}_q}^j - \mathbf{g}_{\mathbf{f}_{s_p}}^j\rVert_2
    \end{split},
\end{align}
for all $n = s_1, s_2, ..., s_P$ with the model parameter $\phi$. $d(\cdot,\cdot)$ refers to the similarity distance between the two comparing instances. $d\langle\cdot, \cdot\rangle$ is the actual metric function to calculate the total similarity aggregation between their global representation, which is the sum of Euclidean distances between every pair of aligned comparing unit $\mathbf{g}_{\mathbf{f}_q}^j$ and $\mathbf{g}_{\mathbf{f}_{s_p}}^j$. During training, the model optimizes $\phi$ by a negative log-probability loss of $\mathcal{L}_\phi = -\log p_\phi(y_{q_{\mathit{train}}}=\mathrm{y}_{q_{\mathit{train}}}|\mathbf{x}_{q_{\mathit{train}}})$ from the predicted probability for the true class $\mathrm{y}_{q_{\mathit{train}}}\in \{s_{1_{\mathit{train}}}, ..., s_{P_{\mathit{train}}}\}$ of $\mathbf{x}_{q_{\mathit{train}}}$. During testing, having a trained model parameter $\phi '$, the class $s_{\mathit{pred}}$ which meets $p_{\phi '}(y_{q_{\mathit{test}}}=s_{\mathit{pred}} | \mathbf{x}_{q_{\mathit{test}}}) = \max\{p_{\phi '}(y_{q_{\mathit{test}}}=s_n|\mathbf{x}_{q_{\mathit{test}}}) | n\in \{s_{1_{\mathit{test}}}, ..., s_{P_{\mathit{test}}}\}\}$ will become the predicted class for $\mathbf{x}_{q_{\mathit{test}}}$. 

\subsection{Adaptive Dependency Learning (ADL)}\label{subsec:adl}
\begin{figure}[t]
\centering
\includegraphics[width=\columnwidth]{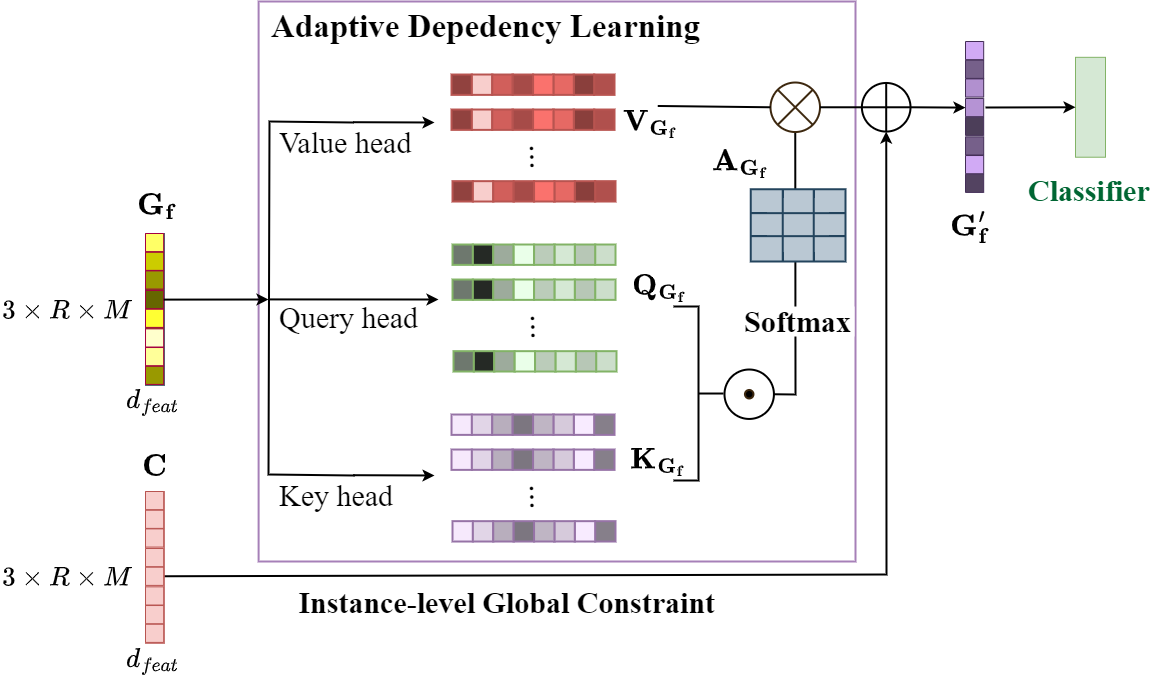}
\caption{ADL module learns to adaptively distribute contextual comparing focus for each unit on action-related local embedding content. }
\label{fig:adl}
\end{figure}
The classification up till now determines the total similarity by an unbiased sum of embedding distances from all comparing units. It equalizes the decision impact of local comparison from action-critical/noisy sub-areas, which hampers the generalization process of correct classification. To emphasize the learning reliance on the former and avoid the negative learning from the latter, we design a self-attention-based module ADL to distribute the contextual significance for each comparing unit. As shown in Fig. \ref{fig:adl}, we prepare three parametric matrices, known as the value head $\mathbf{V}:\mathbb{R}^{d_{\mathit{feat}}} \times \mathbb{R}^{d_{\mathit{feat}}}$, the key head $\mathbf{K}:\mathbb{R}^{d_{\mathit{emb}}} \times \mathbb{R}^{d_{\mathit{feat}}}$ and the query head $\mathbf{Q}:\mathbb{R}^{d_{\mathit{emb}}} \times \mathbb{R}^{d_{\mathit{feat}}}$, where $d_{\mathit{emb}}$ is the output embedding size. After calculating $\mathbf{K}_{\mathbf{G}_\mathbf{f}} = \mathbf{K}\cdot\mathbf{G}_\mathbf{f}$, $\mathbf{Q}_{\mathbf{G}_\mathbf{f}} = \mathbf{Q}\cdot\mathbf{G}_\mathbf{f}$ and $\mathbf{V}_{\mathbf{G}_\mathbf{f}} = \mathbf{V}\cdot\mathbf{G}_\mathbf{f}$ for a global representation $\mathbf{G}_\mathbf{f}$, we generate a matrix of normalized attention scores $\mathbf{A}_{\mathbf{G}_\mathbf{f}}$ which captures the action-based contextual dependency for each unit on every other unit in $\mathbf{G}_\mathbf{f}$. Using the scores as the weights, the new content of a comparing unit $\mathbf{g}'_\mathbf{f}$ is a weighted sum of the value-head output from its original embedding in $\mathbf{g}_\mathbf{f}$ and every other comparing unit. To express this as matrix-level operations:
\begin{align} \label{eq:attention}
\mathbf{A}_{\mathbf{G}_\mathbf{f}} &= \frac{\exp((\mathbf{K}_{\mathbf{G}_\mathbf{f}}\cdot \mathbf{Q}_{\mathbf{G}_\mathbf{f}})/\sqrt{d_{\mathit{emb}}})}{\sum\nolimits_{j=1}^{3\times R\times M}\exp((\mathbf{K}_{\mathbf{G}_\mathbf{f}}\cdot \mathbf{Q}_{\mathbf{G}_\mathbf{f}})/\sqrt{d_{\mathit{emb}}})}\ ,\\
\mathbf{G'}_\mathbf{f} &= \mathbf{A}_{\mathbf{G}_\mathbf{f}} \mathbf{V}_{\mathbf{G}_\mathbf{f}} + \mathbf{C}\ .
\end{align}

Pure local embedding representation clusters weakly for intra-class samples. Thus we remain the global average embedding $\mathbf{f}_{\mathit{glob}} = \frac{1}{T_{\mathit{feat}}\times U\times M}\sum_{t=1}^{T_{\mathit{feat}}}\sum_{v=1}^{U}\sum_{m=1}^{M}\mathbf{f}_{vt}^m$ from all body joints and temporal feature dimensions as a simple instance-level constraint and add it into each unit by an expansion matrix $\mathbf{C}$. We now jointly train the attention matrices and the feature encoder together in an end-to-end manner. Eventually, when figuring the new embedding for each unit, the original features from action-critical units would not only be persisted in their units but also be transmitted to other units as high-attention contextual supplements because they contain more invariant information for correct classification. On the other hand, the new embedding for the units whose original features are low-attention (i.e. noisy) would suppress their old information and be more amended by the contextual features from their correlated high-attention units or global embedding. Eventually, this promotes a targeted learning direction that emphasizes the decision weights of similarity measurements according to the native and contextual features from action-critical units, and suppresses the impacts from action-noisy units.  

\section{Experiments}
\label{sec:evaluation}
Aligning to \cite{skl-dml}, we evaluate our model on the \textit{NTU-RGB+D 120} dataset \cite{3} which provides large-scale action recognition scenarios. According to its official protocol \cite{3}, the dataset is split into an 100-class auxiliary set and a 20-class evaluation set with non-overlapping classes, and each class in the evaluation set has only one reference sample. Our experiments are developed in two stages. One is the standard performance examination based on its one-shot testing protocol, checking the model's general performance trained from the full auxiliary set and its corresponding learning efficiency under different reduced auxiliary sizes. We compare our outcomes with the results in all previous related papers and analyze the difference between them and our model. Secondly, we carry out the ablation study to determine the exact learning effect brought by each model component. 
\subsection{Dataset and Evaluation Protocol}
The \textit{NTU-RGB+D 120} \cite{3} dataset is a large action recognition dataset that contains 114,480 skeleton sequences of 120 action classes from 106 subjects in 155 different camera views. The action labels range from daily/health-related individual or mutual actions. Obtained by Kinect depth sensors, each sequence provides real-world 3-d coordinates of 25 body joints for each skeleton (of up to 2 attending performers). Our model needs to first get trained on the available auxiliary set to provide a general common embedding space for any newly coming action class. During the testing stage, our model predicts the evaluating samples by finding their nearest class reference neighbors according to the local-component-based comparison between their embedding representation. For the general performance examination, we use the whole 100-class auxiliary set to train our model. For the auxiliary reduction experiment, aligning to the benchmarks in \cite{3}, we apply a variable control on the auxiliary class size in a range of 20, 40, 60, 80, and 100. For the ablation study, we maintain the same experiment settings under different auxiliary sizes but apply them to the different versions of our model with respective variable control on each specific component.
\subsection{Implementation Details}
\label{subsec:implements}
The model is implemented in PyTorch \cite{4}. To unify the sequence temporal length, we apply an average-frame-sampling/zero-padding for the skeleton sequences longer/shorter than 75 frames (the mode value for the distribution from all original lengths). For the frontal-viewing pre-process, we borrow the algorithm from \cite{skeleton-adjust} and regard the first actor's facing direction in the first frame as the standard frontal direction to the camera throughout a sequence. Concretely, the facing is calculated as the orthogonal direction for the direction from the skeleton's left hip to its right hip and the direction from its central hip to its spine. Then the 3-d location of every body joint in all the frames is vertically rotated to transform to the coordinate system under the new viewing angle. Apart from the convolution sampling strategy, our feature encoder is aligned to \cite{2}, composed of 10 iterative blocks of spatial and temporal convolution layers. For each spatial convolution layer, $L$ is set to be 1. The output dimension for each block evolves as $64\times 4\rightarrow 128\times 3\rightarrow 256\times 3$. The embedding dimension in ADL is 256. We conduct each experiment with a maximum training of 100 epochs on 2 NVIDIA P100 GPUs, and apply early stops when the validating accuracy doesn't improve in the latest 10 epochs. An Adam optimizer and cosine annealing are used to schedule the learning rate with a starting value of $10^{-3}$ and the weight decay of $10^{-6}$. During training, we mainly adopt the episodic learning algorithm (see Section \ref{subsec:global-repres}), in which each training-use sub-task has the same 20-way-1-shot setting from the testing protocol. As a controlled experiment, we also attempted training our model under a traditional style, in which the model is trained by normal batch learning with a batch size of 64. For the encoded feature $\mathbf{f}$ of an input example, we performed a global average pooling on $\mathbf{G}'_\mathbf{f}$ to get a 256-dimension feature vector and then connected it with a SoftMax classifier to train the model by the standard cross-entropy loss. During testing, we disconnected the classifier and used the trained encoder, embedder and ADL to perform the same nearest-neighbor-based classification as episodic learning. 

\subsection{Results}
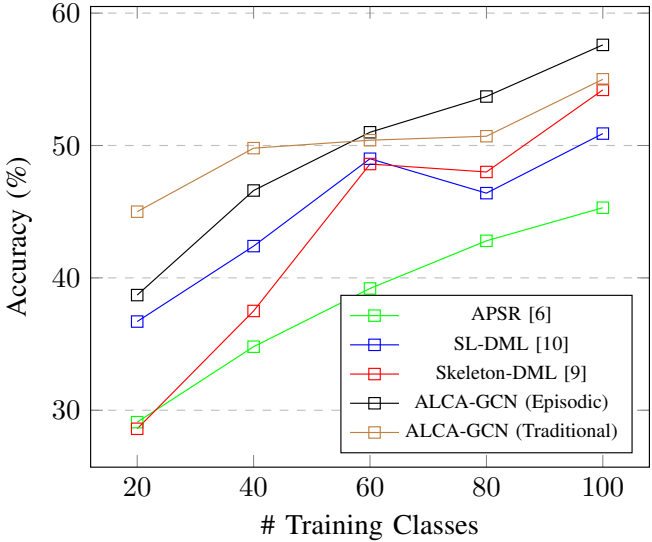
\begin{figure}[!ht]
\centering
\resizebox{\columnwidth}{!}{
\begin{tikzpicture}
\begin{axis}[
    xlabel=\# Training Classes, 
    ylabel=Accuracy (\%), 
    xtick={20, 40, 60, 80, 100},
    tick align=inside,
    legend style={font=\fontsize{7}{8}\selectfont},
    legend pos=south east,
    ymajorgrids=true,
    grid style=dashed
    ]

\addplot[sharp plot,mark=square,green] plot coordinates { 
    (20,29.1)
    (40,34.8)
    (60,39.2)
    (80,42.8)
    (100,45.3)
};
\addlegendentry{APSR \cite{3}}

\addplot[sharp plot,mark=square,blue] plot coordinates {
    (20,36.7)
    (40,42.4)
    (60,49.0)
    (80,46.4)
    (100,50.9)
};
\addlegendentry{SL-DML \cite{sl-dml}}

\addplot[sharp plot,mark=square,red] plot coordinates {
    (20, 28.6)
    (40, 37.5)
    (60, 48.6)
    (80, 48.0)
    (100, 54.2)
};
\addlegendentry{Skeleton-DML \cite{skl-dml}}

\addplot[sharp plot,mark=square,black] plot coordinates {
    (20, 38.7)
    (40, 46.6)
    (60, 51.0)
    (80, 53.7)
    (100, 57.6)
};
\addlegendentry{ALCA-GCN (Episodic)}

\addplot[sharp plot,mark=square,brown] plot coordinates {
    (20, 45.0)
    (40, 49.8)
    (60, 50.4)
    (80, 50.7)
    (100, 55.0)
};
\addlegendentry{ALCA-GCN (Traditional)}
\end{axis}
\end{tikzpicture}
}
\caption{Visualized 1-shot accuracy variation on \textit{NTU-RGB+D 120} with different auxiliary training set sizes.}
\label{fig:line-chart}
\end{figure}

\begin{table}
\centering
\resizebox{.6\columnwidth}{!}{%
\begin{tabular}{cc}
\hline  
Approach &Accuracy\\
\hline
Attention Network \cite{attention-network} &41.0\\
Fully Connected \cite{attention-network} &42.1\\
Average Pooling \cite{average-pooling} &42.9\\
APSR \cite{3} &45.3\\
TCN \cite{tcn} &46.5\\
SL-DML \cite{sl-dml} &50.9\\
Skeleton-DML \cite{skl-dml} &54.2\\
ALCA-GCN (Episodic) &\textbf{57.6}\\
ALCA-GCN (Traditional) &\textbf{55.0}\\
\hline  
\end{tabular}
}
\caption{General 1-shot action recognition results (\%) on \textit{NTU-RGB+D 120} with full training on all 100 auxiliary classes.}
\label{tab:general}
\end{table}
\textbf{General and Training Set Size Reduction}. 
Table \ref{tab:general} presents our model's general performance for the given 1-shot task, compared to the available solution results in \cite{3,skl-dml,sl-dml,tcn,attention-network,average-pooling} under the same testing protocol. Table \ref{tab:reduction} and Fig. \ref{fig:line-chart} present our model's corresponding learning efficiency under different auxiliary sizes, compared to the available results in \cite{3,skl-dml,sl-dml}. The solutions in \cite{3,tcn,attention-network,average-pooling} all use certain global average embedding for similarity comparison, while \cite{skl-dml,sl-dml} transform skeleton sequences into signal images. The outcomes show that our model learned by either training strategy always performs better than the existing solutions under any auxiliary condition. Concretely, our model trained by traditional learning outperforms the previous state-of-the-art in \cite{sl-dml} with a margin of 8.3\% and 7.4\% for the auxiliary size of 20 and 40, and our model trained by episodic learning outperforms \cite{skl-dml,sl-dml} by a margin of 2.0\%, 5.7\%, 3.4\% for the auxiliary size of 60, 80 and 100. We find that traditional training provides more efficient embedding learning for our model under low auxiliary supports probably because at this moment the training already simulates a similar learning sample distribution to the evaluation task (a 20-way classification). Meanwhile, a larger batch training than episodic learning helps our model more easily get out of the local minimum and find the optimal parameters. On the other hand, episodic learning presents a more stable learning increase for the model's generalized embedding ability by meta-learning from the gradually abundant auxiliary classes. Observing the visualized learning progress under different auxiliary sizes in Fig. \ref{fig:line-chart}, we see that both the global-embedding-based method \cite{3} and our model under traditional training reduce their accuracy improvement speed when the auxiliary size raises from 40 to 60 classes or 60 to 80 classes. More seriously, \cite{skl-dml,sl-dml} face temporary learning confusion when the auxiliary size raises from 60 classes to 80 classes, having a 2.6\% and 0.6\% accuracy drop. Contrastly, our model under episodic learning demonstrates a steady learning increase, enlarging the advantage gap when the auxiliary size is 80 or 100 classes. 

\begin{table}
\centering
\resizebox{\columnwidth}{!}{%
\begin{tabular}{c|c|c|c|c|c}
\hline  
\# Training Classes &20 &40 &60 &80 &100 \\
\hline  
APSR \cite{3} &29.1 &34.8 &39.2 &42.8 &45.3\\
SL-DML \cite{sl-dml} &36.7 &42.4 &49.0 &46.4 &50.9\\
Skeleton-DML \cite{skl-dml} &28.6 &37.5 &48.6 &48.0 &54.2\\
ALCA-GCN (Episodic) &38.7 &46.6 &\textbf{51.0} &\textbf{53.7} &\textbf{57.6}\\
ALCA-GCN (Traditional) & \textbf{45.0} &\textbf{49.8} &50.4 &50.7 &55.0\\
\hline  
\end{tabular}
}
\caption{1-shot action recognition results (\%) on \textit{NTU-RGB+D 120} with different auxiliary training set sizes.}
\label{tab:reduction}
\end{table}

\textbf{Ablation Study}. Table \ref{tab:ablation} records the ablation study results on the detailed learning effect brought by each model component (using episodic learning under the same configuration as the full training). We separate the research objects into three types of components: \textbf{convolution sampling strategy}, \textbf{comparing unit division}, and \textbf{instance-level constraints}. For convolution sampling, we examine the influence of spatial feature extraction under different scopes by only using the original convolution scheme in \cite{2} or our body-part-based scheme. For comparing unit division, we consider the learning efficiency under different similarity metrics, including conducting the measurements by global average embedding (labeled as None because there is no local division), pure spatial-wise comparing units (dividing temporally-averaged features according to body-part partitions), or pure temporal-wise comparing units (dividing body-joint-averaged features according to temporal sections). For instance-level constraints, the study examines the performance drop when the ADL module or the global embedding constraint is removed from the original ALCA-GCN.  

The outcome indicates that the overall best result under any auxiliary condition is achieved by the full ALCA-GCN. For the spatial convolution scheme, the visual features collected from either skeleton-based or body-part-based neighbor sampling provide comparable distinction validity for classification. The full ALCA-GCN concatenates them to provide a more comprehensive feature description and improves by 1.0\%, 2.3\%, 4.3\%, and 4.2\% for the auxiliary size of 40, 60, 80, and 100. A similar situation also appears for dividing comparing units only on spatial or temporal dimensions. We observe that using global average embedding predicts better than using single-dimensional comparing units by 5.4\%, 0.8\%, and 1.4\% when the auxiliary size is 40, 80, and 100. But using double-dimensional comparing units in the full ALCA-GCN outperforms using global embedding under every condition with a respective advantage of 3.8\%, 0.8\%, 4.8\%, 2.6\%, 3.3\%. Finally, both the ADL module and global embedding constraints are verified as positive regulations for our similarity metric. Especially, the performance boost brought by ADL is the most obvious. Except for the situation under 20 auxiliary classes, it steadily provides an increase of 4.4\%, 4.4\%, 4.8\%, and 7.0\% when the auxiliary size grows to 40, 60, 80, and 100 classes. Under smaller auxiliary sizes, the models with only body-part-based spatial convolution or without ADL could achieve similar learning results to the full model, because the embedding discrimination for only 20 training classes is relatively easy. With more abundant and complicated auxiliary classes, the model needs to develop its generalized embedding ability with more explicit and refined pattern recognition on potential class-specific movements, in which our ADL contributes significantly by filtering action-related local features.

\begin{table}
\centering
\resizebox{\columnwidth}{!}{%
\begin{tabular}{c|c|c|c|c|c|c}
\hline
 &\# Training Classes &20 &40 &60 &80 &100 \\
\hhline{=|=|=|=|=|=|=}
\multirow{2}{*}{Sampling strategy} & Body-part-based &\textbf{38.7} &45.6 &45.7 &49.2 &53.4\\
 & Skeleton-based &37.5 &43.3 &48.7 &49.4 &51.6\\
\hline
\multirow{3}{*}{Division strategy} & None &34.9 &45.8 &46.2 &51.1 &54.3 \\
 & Spatial-wise &31.4 &40.4 &47.4 &50.3 &52.9\\
 & Temporal-wise &35.5 &40.2 &44.2 &46.5 &50.0\\
\hline
\multirow{2}{*}{Constraints} & Without ADL &38.6 &42.2 &46.6 &48.9 &50.6 \\
 & Without global constraints &31.3 &45.8 &47.1 &51.6 &55.1\\
\hline 
\multicolumn{2}{c|}{ALCA-GCN} &\textbf{38.7} &\textbf{46.6} &\textbf{51.0} &\textbf{53.7} &\textbf{57.6}\\
 \hline
\end{tabular}
}
\caption{Ablation study (\%) on \textit{NTU-RGB+D 120} for each component in our proposed model. }
\label{tab:ablation}
\end{table}

\section{Conclusion}
\label{sec:conclusion}
In this paper, we suggest a novel metric-based few-shot learning for skeleton-based action recognition by decomposing the similarity comparison to an adaptive sum of embedding distance measurements on local spatial-temporal areas. Our encoder convolutionally extracts the features under the hierarchical receptive fields, then our embedder generates independent local comparing units. To emphasize/suppress the feature distinction learning on action-related/noisy units, our ADL module adaptively adjusts each unit's measurement impact according to its instance-level attention with global constraints. We examined our model's general performance and learning effect brought by each component under an extensive experiment setup. The results proved that our model outperforms global-embedding-based and signal-based methods by providing a more instantaneously action-representative embedding for the comparison between limited support and query samples. Using episodic learning, the model could steadily develop its embedding ability by meta-learning from the increased auxiliary resources while previous methods could face a generalization bottleneck. Our solution revealed that there exist unfully-explored physical properties in skeleton sequences that are different from RGB-based characteristics but have invariant structural meanings to facilitate skeleton-specific learning. Under few-shot scenarios, combining our solution with more scalable structure learning for feature representation is a promising and interesting direction to explore in the future research.  

\bibliographystyle{./IEEEtran}
\bibliography{references}


\end{document}